\documentclass{article}

\usepackage[numbers]{natbib}
\usepackage[final]{nips_2017}

\usepackage[utf8]{inputenc} 
\usepackage[T1]{fontenc}    
\usepackage{hyperref}       
\usepackage{url}            
\usepackage{booktabs}       
\usepackage{amsfonts}       
\usepackage{nicefrac}       
\usepackage{microtype}      
\usepackage{amsmath,graphicx}
\usepackage{amsfonts}
\usepackage{float}
\usepackage{booktabs}

\title{Multi-Domain Adversarial Learning for Slot Filling in Spoken Language Understanding}

\author{
	Bing Liu$^1$, Ian Lane$^1$$^,$$^2$ \\
    $^1$Electrical and Computer Engineering, Carnegie Mellon University \\
    $^2$Language Technologies Institute, Carnegie Mellon University \\
    \texttt \{liubing, lane\}@cmu.edu
}

\begin{document}

\maketitle

\begin{abstract}
    The goal of this paper is to learn cross-domain representations for slot filling task in spoken language understanding (SLU). Most of the recently published SLU models are domain-specific ones that work on individual task domains. Annotating data for each individual task domain is both financially costly and non-scalable. In this work, we propose an adversarial training method in learning common features and representations that can be shared across multiple domains. Model that produces such shared representations can be combined with models trained on individual domain SLU data to reduce the amount of training samples required for developing a new domain. In our experiments using data sets from multiple domains, we show that adversarial training helps in learning better domain-general SLU models, leading to improved slot filling F1 scores. We further show that applying adversarial learning on domain-general model also helps in achieving higher slot filling performance when the model is jointly optimized with domain-specific models. 
\end{abstract}

\section{Introduction}
    Spoken language understanding (SLU) system is a critical component in today's task-oriented conversational agents. A typical pipeline of SLU system is to first classify the domain of a user's query, and then identify user's intent and extract semantic frames from the user's natural language utterance. These three tasks are usually referred to as domain classification, intent identification, and slot filling. Many statistical learning methods have been developed for these tasks over the past years. Domain classification and intent detection can be treated as sequence classification problems, and classifiers like support vector machines and deep neural network \cite{sarikaya2011deep,kim2014convolutional} can be applied. Slot filling can be framed as a sequence tagging task. Popular methods for solving sequence tagging include maximum entropy Markov models (MEMMs) \cite{mccallum2000maximum}, conditional random fields (CRFs) \cite{raymond2007generative}, and recurrent neural networks (RNNs) \cite{mesnil2015using,Liu+2016,chen2016end}. 
    
    Many of the previously proposed SLU models focus on solving tasks in individual domains. Annotating semantic tags for each individual task domain is both financially costly and non-scalable. In this work, we focus on learning common knowledge and representations that can be shared across multiple domains for slot filling in SLU. This shared knowledge can help domain-specific SLU model trained with limited in-domain data to reduce the amount of annotated data required for developing a new domain. We want to train a domain-general model that focuses only on learning common patterns across multiple domains, and jointly optimize it with a domain-specific model that preserves in-domain knowledge for slot filling. 

    A straightforward manner to learn cross-domain representations is to train a single model on a combination of the data from all domains. Such model, however, may still learn disjoint domain-specific features due to the very different data distributions in different domains. To enforce the model to discover features and structures that are common across multiple domains, we propose to apply domain adversarial learning in training the domain-general model. The intuition behind is that good common representations across domains are the ones that with which a system cannot recover the domain of the original inputs \cite{ben2010theory}. Once obtaining such a domain-general model, we ensemble it with domain-specific models for best slot filling performance in each task domain. We show in the experiment results that domain-general model trained on all source data can benefit from domain adversarial learning in achieving better slot filling F1 scores. Adversarial learning on domain-general model also brings benefit when the model is jointly optimized with domain-specific models. 
    
    The remainder of the paper is organized as follows. In section 2, we discuss related work on multi-domain slot filling. In section 3, we present our proposed model and learning method. In Section 4, we discuss the experiment setup and the results. In section 5 we conclude the work. 
    
\section{Related Work}
\label{sec:related_work}
    Multi-domain spoken language understanding has been actively studied in literature. Hakkani-Tur et al. \cite{hakkani2016multi} proposed a multi-domain SLU model using Bi-directional LSTM. They train the multi-domain model by simply using data from all the domains and let the data from each domain to reinforce each other. Similarly, Jaech et al. \cite{jaech2016domain} designed an LSTM based multi-domain model that has shared word embeddings and LSTM parameters across domains and separated softmax layer for each domain. The word embeddings and LSTM parameters capture the across-domain knowledge. In our proposed model, we introduce an additional domain adversarial loss in training domain-general model using data from all sources. This enforces the model to learn common features that are general across multiple domains. 
    
    Another line of research that is closely related to our work is domain adaptation for language understanding. Kim et al. \cite{kim2016frustratingly} proposed a neural generalization of the feature augmentation domain adaptation method \cite{daumeiii:2007:ACLMain}. Their model uses an LSTM to captures global patterns by training on data across all domains, and adapt it to one domain with another LSTM trained on only in-domain data. Our proposed method differs in how we train the domain-general model by introducing domain adversarial loss. Adversarial learning has also been applied in domain adaptation. Zhang et al. \cite{zhang2017aspect} used adversarial training to learn domain-invariant features for document classification task. In their model, an adversarial domain classification loss is subtracted directly from the overall system loss function. All model parameters are thus discouraged to make correct domain prediction. In our method, we enforce the domain classifier to take the best classification efforts, and only discourage the underlying network parameters to make correct domain prediction so as to force them to learn domain-invariant features. This is closer to the training of generative adversarial networks \cite{goodfellow2014generative}.

\section{Proposed Method}
\label{sec:method}

\subsection{Bi-LSTM Slot Filling Model}
    Let $\mathbf{w} = (w_{1}, ..., w_{T})$ represent an utterance with $T$ words, and let $\mathbf{y} = (y_{1}, ..., y_{T})$ represent the aligned slot label sequence with the same length as the input word sequence. We use a bidirectional LSTM (bi-LSTM) to encode the sequence of words to continuous representations. 
    The bi-LSTM encoder outputs at each step $t$ is obtained by concatenating the forward and backward LSTM state outputs: $h_{i} = [\overrightarrow{h_{i}}, \overleftarrow{h_{i}}]$. Based on the bi-LSTM encoder output at each step $t$, a probability distribution $P(y_{t})$ over all slot labels for the word input is produced:
            \begin{equation}
                P(y_{t} \hspace{0.5mm} | \hspace{0.5mm} \mathbf{w}; \theta _{y}) = \operatorname{SlotLabelDist}(h_{t})
            \end{equation}
    where $\operatorname{SlotLabelDist}$ is a multi-layer perceptron (MLP) parameterized by $\theta _{y}$ with $\operatorname{softmax}$ outputs over all slot label candidates. We train the bi-LSTM slot filling model by optimizing the parameters $\theta$ to minimize the cross-entropy loss $L_y$ of the predicted and true distributions for slot label at each step:
            \begin{equation}
                \min_{\theta _{s}, \theta _{y}} {L_y} = \min_{\theta _{s}, \theta _{y}} - \frac{1}{T} \sum_{t=0}^{T} \log P(y_t^* | \mathbf{w}; \theta _{s}, \theta _{y}) 
            \end{equation}
    where $y_t^*$ is the ground truth slot label at step $t$, and $\theta _{s}$ represents the parameters of the bi-LSTM and the word embeddings.

\subsection{Domain Adversarial Learning}
    We collect annotated data sets in multiple task domains, and train slot filling models for each domain using only in-domain data. 
    We further train a domain-general model using a combination of the data from all sources. To enforce the model to learn domain-invariant features, we apply adversarial learning when training the domain-general model. 
    We introduce a domain classifier whose job is to identify the domain of the input utterance by taking in the representations from the bi-LSTM state outputs. We train the domain classifier to minimize the domain classification loss. On the other hand, we update the parameters of the network underlying the domain classifier to maximize the domain classification loss, which works adversarially towards the domain classifier. We thus expect the model to learn features and structures that are general across domains. 

\begin{figure*}[t]
    \centering
    \includegraphics[width=360px]{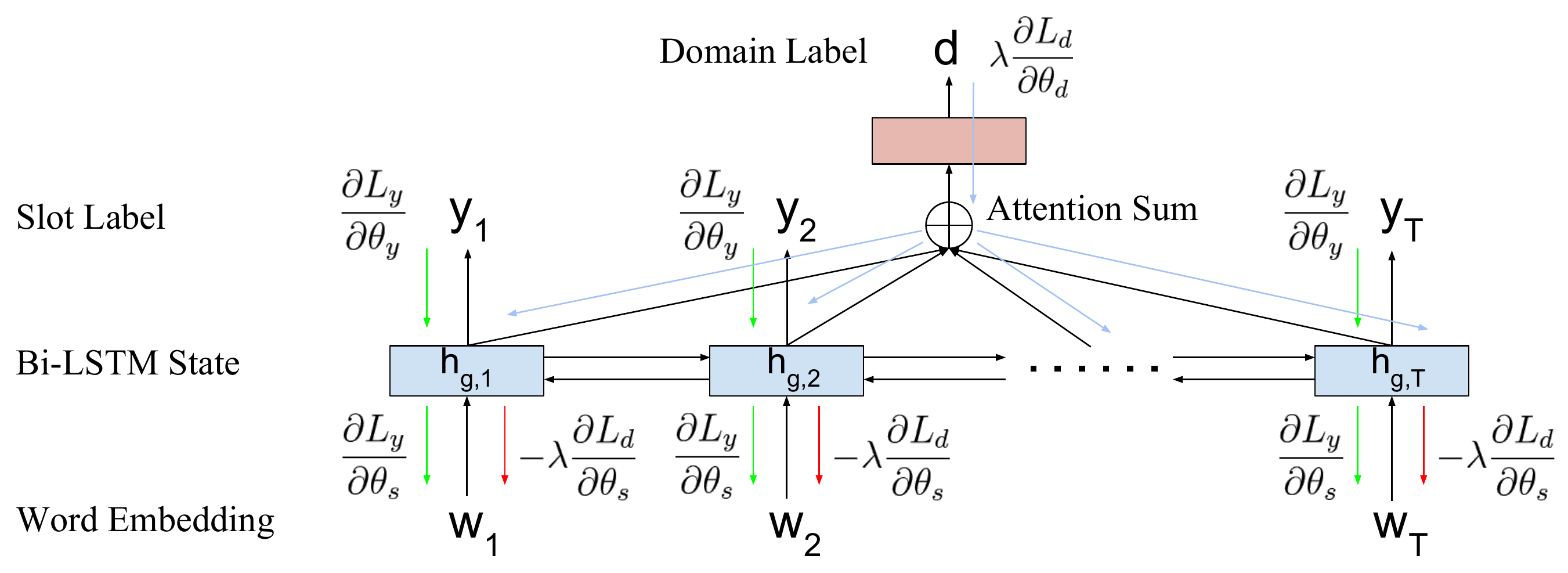}
    \caption{{ Domain-general slot filling model with adversarial learning. }}
    \label{fig:domain_general_rnn}
\end{figure*} 

\textbf{Attention Domain Classifier} \hspace{3mm} The domain classifier takes in the representation of an utterance and estimate its domain. A probability distribution $P(d)$ over all domain labels is generated by:
            \begin{equation}
                P(d \hspace{0.5mm} | \hspace{0.5mm} \mathbf{w}; \theta _{d}) = \operatorname{DomainLabelDist}(c)
            \end{equation}
where $c$ is an attention sum over encoder state outputs $(h_{1}, ..., h_{T})$:
        \begin{equation}
            c = \sum_{t=1}^{T}\alpha_{t}h_{t}
        \end{equation}        
        and
        \begin{equation}
        \begin{split}
            \alpha_{t} = \frac{\operatorname{exp}(e_{t})}{\sum_{k=1}^T\operatorname{exp}(e_{k})}, \hspace{3mm} e_{t} = g(h_{t})
        \end{split}
        \end{equation} 
        $g$ a feed-forward neural network with a single output node. 

\textbf{Adversarial Training} \hspace{3mm}  We train the domain classifier by optimizing its parameter set $\theta _d$ to minimize the cross-entropy loss $L_d$ of the predicted and true domain class distributions:
        \begin{equation}
            \max_{\theta _{s}} \min_{\theta _{d}} {L_d} = \max_{\theta _{s}} \min_{\theta _{d}} - \log P(d^* | \mathbf{w}; \theta _{s}, \theta _{d}) 
        \end{equation}
where $d^*$ is the ground truth domain class label. 

In updating the network parameters $\theta _s$ underlying the domain classifier, we reverse the gradient $\frac{\partial L_d}{\partial \theta_s}$ during back-propagation with the aim to force the model to produce utterance representations that maximize domain classification confusion. This operation is performed by connecting the domain classifier with underlying network via a gradient reversal layer \cite{ganin2016domain}. The overall loss function for slot filling and adversarial domain classification is then calculated by:
        \begin{equation}
            L = L _y  + \lambda L _d
        \end{equation}
where $\lambda$ is the coefficient for the adversarial domain classification loss.

\subsection{Joint Optimization}
\begin{figure*}[t]
    \centering
    \includegraphics[width=320px]{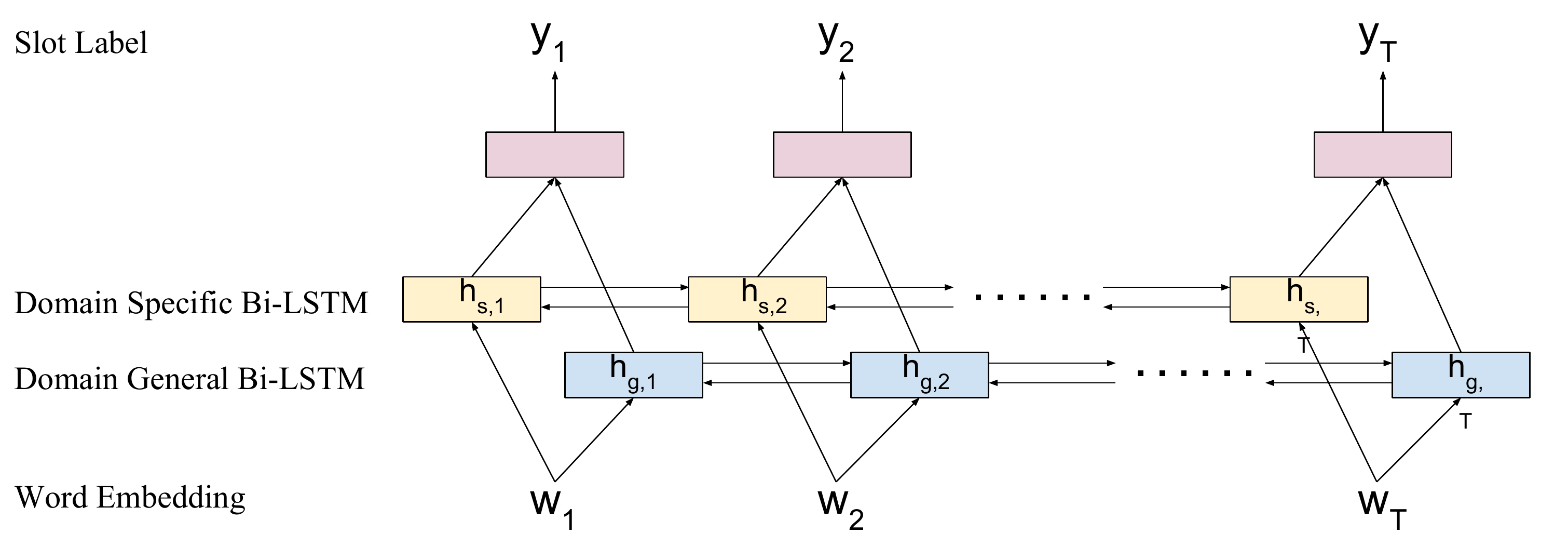}
    \caption{{ Joint domain-specific and domain-general slot filling model. }}
    \label{fig:ensemble_rnn}
\end{figure*}

Figure \ref{fig:ensemble_rnn} shows the network architecture for the joint domain-specific and domain-general slot filling model. At the input layer, embedding of a word is connected to both domain-specific and domain-general Bi-LSTMs. Outputs of the two models are fed to an MLP. A softmax function on top of the MLP produces the probability distribution over all slot labels. We pre-train the domain-specific and domain-general Bi-LSTMs separately. During joint optimization, we fix the parameters of the two Bi-LSTMs, and only optimize the MLP parameters in the output layer.

\section{Experiments}
\subsection{Data Sets}
    We evaluate the proposed model and learning methods on data sets across multiple domains. Each data set consists of semantically tagged training and test corpus in standard BIO format. Table \ref{tab:data_stats} shows the data set statistics. 
    
    \textbf{ATIS} \hspace{3mm} Airline Travel Information Systems (ATIS) data set is widely used in SLU research. It consists of spoken queries on flight related information. 
    
    \textbf{MIT Restaurant Corpus}\footnote{The MIT semantic tagging corpus can be downloaded from: https://groups.csail.mit.edu/sls/downloads} \hspace{3mm} The MIT restaurant data set consists of spoken queries on restaurant searching and booking.
    
    \textbf{MIT Movie Corpus} \hspace{3mm} The MIT movie data set consists of two sets of semantically tagged corpora. The eng corpus contains simple queries, and the trivia10k13 corpus contains more complex queries. There are a few shared slot labels between these two corpora. 

    \begin{table}[th]
      \caption{ Statistics of Data Sets. }
      \label{tab:data_stats}
      \centering
      \begin{tabular}{l c c c c c}
        \toprule
        \textbf{}       & \textbf{}         & \textbf{MIT}      & \textbf{MIT Mov.}     & \textbf{MIT Mov.} & \textbf{}\\
        \textbf{Datasets}   & \textbf{ATIS}     & \textbf{Rest.}    & \textbf{eng}          & \textbf{trivia10k13} & \textbf{Combined}\\
        \midrule
        \textbf{Train set size}     & 4978      & 7660    & 9775        & 7816  &   30229\\
        \textbf{Test set size}      & 893       & 1521    & 2443        & 1953  &   6810\\
        \textbf{Vocab size}         & 572       & 4166    & 7481        & 12145 &   16049\\
        \textbf{Slot label size}    & 127       & 17      & 25          & 25    &   191\\
        \bottomrule
      \end{tabular}   
    \end{table}
    
\subsection{Training Settings}
    We set the number of units in LSTM cell as 128. The default LSTM forget gate bias is set to 1. Word embedding size is set as 128. We randomly initialize the word embedding matrix and fine-tune it with other model parameters during model training. Dropout rate 0.5 is applied to the non-recurrent connections during model training for regularization. Maximum norm for gradient clipping is set to 5. We perform mini-batch training (batch size 16) using Adam optimization with initial learning rate of 1e-3.
    
\subsection{Results and Analysis}
    \begin{table}[th]
      \caption{ Slot Filling F1 scores of the proposed method for data sets in different task domains. }
      \label{tab:slot_filling_results}
      \centering
      \begin{tabular}{l c c c c c}
        \toprule
        \textbf{} & \textbf{}  & \textbf{MIT}  & \textbf{MIT Mov.}  & \textbf{MIT Mov.} & \textbf{} \\
        \textbf{Model} & \textbf{ATIS}  & \textbf{Rest.}  & \textbf{eng}  & \textbf{trivia10k13} & \textbf{Comb. } \\
        \midrule
        Deep LSTM \cite{yao2014spoken}                  & 95.08 & -     & -     & -     & - \\
        RNN-EM \cite{peng2015recurrent}                 & 95.25 & -     & -     & -     & - \\
        Encoder-labeler LSTM \cite{kurata-EtAl:2016:EMNLP2016}  & 95.40 & -     & -     & -     & 74.41 \\
        Attention Bi-LSTM \cite{Liu+2016}               & 95.75 & -     & -     & -     & - \\
        BLSTM-LSTM (focus) \cite{zhu2017encoder}        & \textbf{95.79} & -     & -     & -     & - \\
        \midrule
        Dom-Spec                                        & 95.55 & 72.42 & 83.43 & 63.64 & - \\
        \midrule
        Dom-Gen                                         & 94.09 & \textbf{74.25} & 82.95 & 63.34 & 76.03 \\
        Dom-Gen-Adv ($\lambda$=0.01)                    & \textbf{94.51} & 73.87 & \textbf{83.03} & \textbf{63.51} & \textbf{76.55} \\
        Dom-Gen-Adv ($\lambda$=0.1)                     & 93.88 & 73.98 & 82.31 & 62.83 & 76.01 \\
        Dom-Gen-Adv ($\lambda$=1.0)                     & 84.65 & 62.47 & 75.05 & 52.82 & 66.66 \\
        \midrule
        Joint Dom Spec \& Gen                           & 95.62 & \textbf{74.47} & 84.87 & 65.16 & - \\
        Joint Dom Spec \& Gen-Adv ($\lambda$=0.01)      & \textbf{95.63} & 74.23 & \textbf{85.33} & \textbf{65.33} & - \\
        Joint Dom Spec \& Gen-Adv ($\lambda$=0.1)       & 95.52 & 74.36 & 85.32 & 64.95 & - \\
        Joint Dom Spec \& Gen-Adv ($\lambda$=1.0)       & 95.52 & 73.57 & 84.26 & 64.38 & - \\
        \bottomrule
      \end{tabular}   
    \end{table}
    
    Table \ref{tab:slot_filling_results} shows the slot filling evaluation results. We compare the performance of domain-specific models, domain-general models, and joint domain-specific and domain-general models on data sets over multiple domains. For adversarial learning on domain-general models, we compare the model performance by applying different scales of domain adversarial losses. 
    
    We first compare the performance between our domain-specific models and domain-general models. Our domain-specific (Dom-Spec) model achieves near state-of-the-art slot filling F1 score on ATIS comparing to recently published results. Our first domain-general model (Dom-Gen) is trained on a combination of data sets across multiple domains. This Dom-Gen model outperforms Dom-Spec on MIT restaurant data set, but obtains lower F1 scores on ATIS and the two MIT movie data sets. By further applying domain adversarial training (with $\lambda$=0.01), the Dom-Gen-Adv model achieves improved F1 scores on three out of four data sets over the Dom-Gen model. Using larger scale of domain adversarial loss is not beneficial, resulting in decreased F1 scores across all four data sets. Although domain adversarial learning helps the domain-general models to achieve better slot filling F1 scores by learning common features and structures across different domains, their performance is still below that of the domain-specific models. By looking into the prediction mistakes made by domain-general models on ATIS data set, we find that the model assigns label from other domains which is semantically similar but not matched to the ground truth label in ATIS. This label inconsistency issue across multiple domains makes it not feasible to train a general model by simply combining data from different domains.
    
    We further evaluate the model performance by jointly optimizing the domain-specific and domain-general model. Results in Table \ref{tab:slot_filling_results} illustrate that the joint training model consistently outperforms the domain-specific models and the domain-general models. The joint training model achieves larger gain on the MIT movie data set, improving the F1 score by 1.90\% and 1.69\% respectively on the eng corpus and the trivia10k13 corpus over their domain-specific models. Similar to the results in domain-general model evaluation, adversarial learning on domain-general model also brings benefit to the joint model optimization. Joint model training with adversarial learning achieves advanced slot filling performance comparing to non-adversarial joint model training on three out of four data sets.

\section{Conclusions}
    In this work, we propose applying domain adversarial training in learning common features and representations that can be shared among multiple domains for slot filling task in spoken language understanding. Domain-general models producing such shared representations can be combined with domain-specific models to reduce the amount of data required for training SLU model in a new task domain. Experimental results show that domain-general models trained on a combination of data from all domains can benefit from adversarial learning in achieving higher slot filling F1 scores. We further show that applying adversarial training to domain-general model also helps to achieve advanced slot filling performance when the model is jointly optimized with domain-specific models.

\bibliography{sample}
\bibliographystyle{unsrt}

\end{document}